\documentclass[letterpaper, 10 pt, journal, twoside]{IEEEtran}
\ifCLASSINFOpdf
\else
\fi
\usepackage{times}
\usepackage{epsfig}

\usepackage{color} 
\usepackage{amsmath}
\usepackage{amssymb}
\usepackage{graphicx}
\usepackage{multirow}
\usepackage{tabularx}
\usepackage{adjustbox}
\ifCLASSOPTIONcompsoc
 \usepackage[caption=false,font=normalsize,labelfont=sf,textfont=sf]{subfig}
\else
 \usepackage[caption=false,font=footnotesize]{subfig}
\fi
\usepackage{siunitx}
\usepackage{textcomp}

\usepackage{stmaryrd} 
\usepackage{latexsym} 

\usepackage{url}
\usepackage{hyperref} \hypersetup{colorlinks=true, linkcolor=black, urlcolor=RubineRed}
\usepackage{kotex} 

\newcommand{\figref}[1]{Fig.~\ref{#1}}

\newcommand{\tabref}[1]{Tab.~\ref{#1}}
\newcommand{\eqnref}[1]{Eq.~(\ref{#1})}

\newcommand{\secref}[1]{Sec.~\ref{#1}}

\newcommand{\ie}{\textit{i.e.}}
\newcommand{\eg}{\textit{e.g.}}
\newcommand{\etal}{\textit{et al.}}

\usepackage[dvipsnames]{xcolor}

\begin{document}

\title{\LARGE \bf Maximizing Self-supervision from Thermal Image \\ for Effective Self-supervised Learning of Depth and Ego-motion}

\author{Ukcheol Shin, Kyunghyun Lee, Byeong-Uk Lee, and In So Kweon %
\thanks{Manuscript received: January 12, 2022; Revised April, 22, 2021; Accepted June, 1, 2022.
This paper was recommended for publication by Editor Tamim Asfour and C. Cadena Lerma upon evaluation of the Associate Editor and Reviewers’ comments. 
This work was supported by the International Research and Development Program of the National Research Foundation of Korea(NRF) funded by the Ministry of Science and ICT (NRF-2021K1A3A1A21040016).
\textit{(Corresponding author: In So Kweon.)}}
\thanks{U. Shin, K. Lee, B. Lee, and I. S. Kweon are with the School of Electrical Engineering, KAIST, Daejeon 34141, Republic of Korea.
{\tt \{shinwc159, kyunghyun.lee, byeonguk.lee, iskweon77\}@kaist.ac.kr}}%
\thanks{Digital Object Identifier (DOI): see top of this page.}
}

\markboth{IEEE Robotics and Automation Letters. Preprint Version. Accepted June, 2022}
{Shin \MakeLowercase{\textit{et al.}}: Maximizing Self-supervision from Thermal Image for Effective Self-supervised Learning of Depth and Ego-motion} 

\maketitle

\begin{abstract}
Recently, self-supervised learning of depth and ego-motion from thermal images shows strong robustness and reliability under challenging scenarios.
However, the inherent thermal image properties such as weak contrast, blurry edges, and noise hinder to generate effective self-supervision from thermal images.
Therefore, most research relies on additional self-supervision sources such as well-lit RGB images, generative models, and Lidar information.
In this paper, we conduct an in-depth analysis of thermal image characteristics that degenerates self-supervision from thermal images. 
Based on the analysis, we propose an effective thermal image mapping method that significantly increases image information, such as overall structure, contrast, and details, while preserving temporal consistency.
The proposed method shows outperformed depth and pose results than previous state-of-the-art networks without leveraging additional RGB guidance.
\end{abstract}


\IEEEpeerreviewmaketitle

\section{Introduction}

Recently, thermal image based 3D vision applications for a robust robot vision~\cite{choi2018kaist,kim2018multispectral,shin2021self,lu2021alternative} are gradually attracting attention.
Since a thermal camera has a consistent image capturing ability regardless of lighting and weather conditions, it is suitable for robust robot vision.
However, thermal image usually suffers from its inherent image properties, such as low signal-to-noise ratio, low contrast, and blurry edge.
Therefore, lots of researches have been proposed that utilize another heterogeneous sensor, such as RGB camera, depth sensor, IMU, and Lidar, to compensate for the inherent problems.

The aforementioned thermal image properties are a more serious issue for the self-supervised learning of depth and ego-motion estimation.
Therefore, Kim~\etal~\cite{kim2018multispectral} exploits RGB stereo image pair to train thermal image based monocular depth network with spatial RGB image reconstruction loss.
Lu~\etal~\cite{lu2021alternative} utilizes an RGB-to-thermal image translation network for the spatial image reconstruction loss between left thermal-like image and right thermal image to train disparity network.
Lastly, Shin~\etal~\cite{shin2021self} proposed multi-spectral temporal image reconstruction loss that utilizes monocular RGB and thermal video to train single-view depth and multiple-view pose networks. 

\begin{figure}[t]
\begin{center}
\footnotesize
\begin{tabular}{c@{\hskip 0.005\linewidth}c@{\hskip 0.005\linewidth}c@{\hskip 0.005\linewidth}c}
\includegraphics[width=0.32\linewidth]{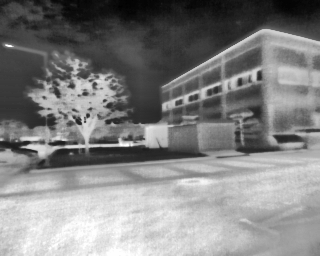} &  
\includegraphics[width=0.32\linewidth]{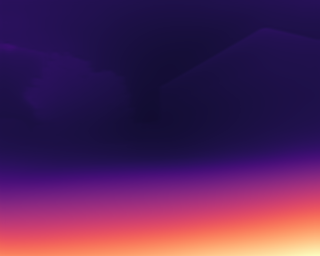} &
\includegraphics[width=0.32\linewidth]{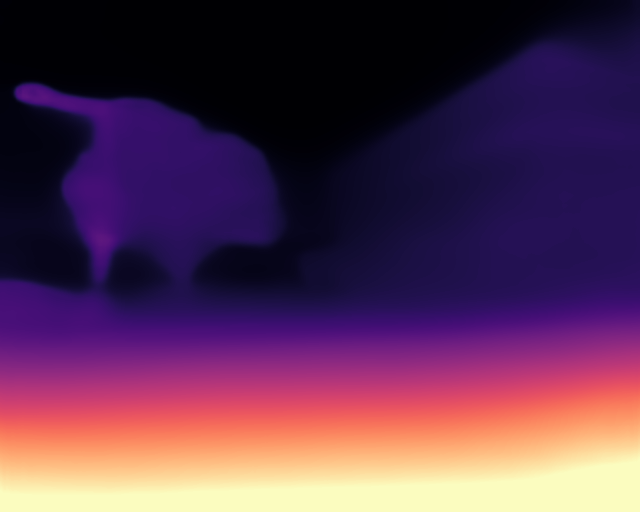} \vspace{-0.03in} \\
\includegraphics[width=0.32\linewidth]{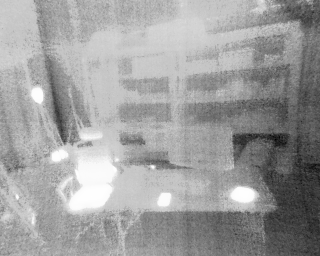} &
\includegraphics[width=0.32\linewidth]{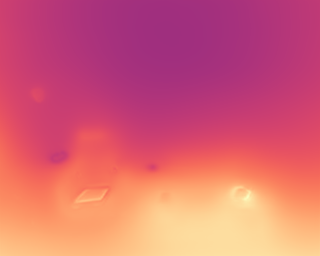} &  
\includegraphics[width=0.32\linewidth]{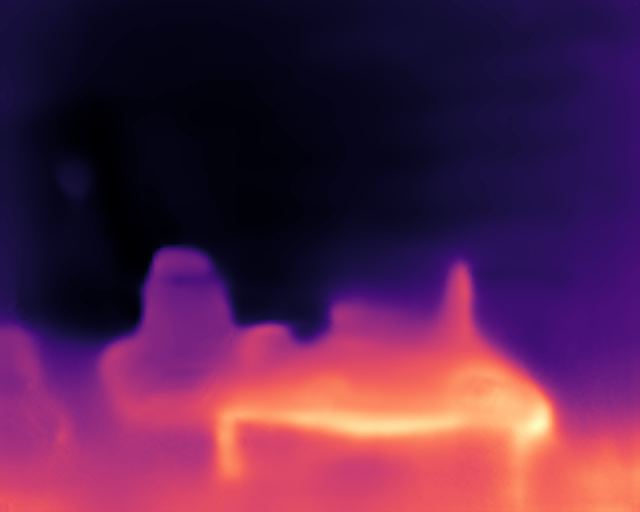} \\
{\bf (a) Thermal image} & {\bf (b) Shin~\etal (T)}  & {\bf (c) Ours}  \\
\end{tabular}
\end{center}
\caption{{\bf Depth comparison between Shin~\etal~\cite{shin2021self} and proposed method}. 
(a) is the temporal consistently re-arranged thermal images with our proposed method, (b) is depth map results trained with Shin~\etal~\cite{shin2021self}, and (c) is the depth map results trained with the our method.
Our proposed temporal consistent temperature mapping method remarkably increases image contrast and details while preserving temporal consistency for effective self-supervisory signal. 
}
\label{fig:teaser}
\end{figure}
However, these methods do not try to solve the fundamental issue of the thermal image and leverage knowledge of additional modalities that usually require careful hardware setting, additional training stage, and accurate calibration process.
In this paper, we first analyze thermal image properties that hinder self-supervised learning of depth and pose networks from thermal images.
After that, we propose an effective thermal image mapping method to train depth and pose networks without any additional modality guidance.
Based on the proposed method, the networks show accurate and sharp depth results in all evaluation datasets, as shown in~\figref{fig:teaser}.

Our contributions can be summarized as follows:
\begin{itemize}
\item We provide in-depth analysis for raw thermal image properties in terms of self-supervised learning perspective. 
Our observation are as follows: 
      \begin{itemize}
        \item{Raw thermal image is suitable choice to preserve temporal consistency.}
        \item{Sparsely distributed thermal radiation values dominate image reconstruction loss.}
        \item{High and low temperature objects degenerate self-supervision from thermal image.}
      \end{itemize}
\item 
Based on the in-depth analysis, we propose a temporal consistent image mapping method that maximizes self-supervision from thermal images by rearranging thermal radiation values and boosting image information while preserving temporal consistency.
\item
We demonstrate that the proposed method shows outperformed results than previous state-of-the-art networks without leveraging additional guidance.
\end{itemize}

\section{Related Works}
\label{sec:related works}
\subsection{Self-supervised Depth and Pose Network}

Self-supervised learning of depth and ego-motion network is the study of 3D understanding that does not rely on expensive Ground-Truth labels.
SfM-learner~\cite{zhou2017unsupervised} is the pioneering work that train neural network to estimate depth and relative pose in a fully unsupervised manner by using temporal image reconstruction. 
Given near-by-view image sequences, one image is warped to the other image coordinate by utilizing an estimated depth map and relative pose. 
The difference between the warped image and the other image is used for a supervisory signal to train depth and pose networks.
However, the violation cases such as brightness inconsistency, non-Labertian surface, homogeneous regions, and moving objects exist in the reconstruction process.
Therefore, invalid pixel masking scheme are variously proposed with the form of explainability mask~\cite{zhou2017unsupervised,vijayanarasimhan2017sfm}, flow-based mask~\cite{yin2018geonet,ranjan2019competitive,wang2019recurrent}, static mask~\cite{godard2019digging}, and depth inconsistent mask~\cite{bian2019unsupervised, bian2021unsupervised}

Feature-map reconstruction loss~\cite{zhan2018unsupervised,shu2020feature} are proposed to complement the self-supervisory signal of the homogeneous regions. 
For the scale-ambiguity problem, several works~\cite{wang2019recurrent,bian2019unsupervised,chen2019self,bian2021unsupervised,kopf2021robust} impose geometric constraints to predict a scale-consistent depth and camera trajectory.
A semantic knowledge also can be leveraged to enhance the feature representation for monocular depth estimation~\cite{chen2019towards,guizilini2020semantically}.
Guizilini~\etal~\cite{guizilini20203d} introduce a detail-preserving representation using 3D convolutions.

In contrast to the researches that focus on the RGB domain, our proposed method focuses on the self-supervised learning of depth and pose estimation from thermal images that has not been actively explored so far.
Furthermore, it is not known whether the above-mentioned techniques can lead to a performance improvement in the thermal spectral domain.
Therefore, based on our empirical observations, we form a concise self-supervised learning method in the thermal spectrum domain.

\subsection{Self-supervised Depth and Pose Network from Thermal Images}

Thermal image based depth or pose estimation networks are one of the crucial topics for a weather and lighting condition invariant 3D understanding.
Kim~\etal~\cite{kim2018multispectral} proposed a self-supervised multi-task learning framework that exploits spatial RGB stereo reconstruction to train single-view depth estimation.
They assume the thermal image is geometrically aligned with the left RGB image. 
They make a synthesized right RGB image with an estimated depth map from the thermal image, left RGB image, and extrinsic parameter.
Lu~\etal~\cite{lu2021alternative} proposed a image translation network based cross-spectral reconstruction loss to train single-view depth network.
Their method requires an RGB stereo pair and one thermal image.
The left RGB image is synthesized to a thermal-like image with the image translation network.
After that, the spatial reconstruction loss between the thermal-like left and real right thermal images is used to train the network.

Shin~\etal~\cite{shin2021self} proposed a multi-spectral temporal consistent loss to train single-view depth and multiple-view pose networks.
For this purpose, they analyze a proper thermal image format for the temporal image reconstruction loss.
They pointed out that a normalized thermal image used in lots of applications~\cite{kim2018multispectral,lu2021alternative,sun2019rtfnet,sun2020fuseseg,kim2021ms} already lost the temporal consistency between the adjacent frames and is not suitable for depth and pose estimation application.
Based on the observation, they proposed a basic form of temperature reconstruction loss that preserve temporal consistency along with a temporal photometric consistency loss. 

However, these methods do not try to solve the fundamental issue of the thermal image and leverage other modalities' information that usually requires careful hardware setting~\cite{kim2018multispectral}, additional training stage~\cite{lu2021alternative}, and accurate extrinsic calibration process~\cite{kim2018multispectral,lu2021alternative,shin2021self}.
In contrast to these studies, our proposed method trains single-view depth and multiple-view pose networks from monocular thermal video without relying on other sensor modalities.
\begin{figure}[t]
\begin{center}
\footnotesize
\begin{tabular}
{c@{\hskip 0.005\linewidth}c@{\hskip 0.005\linewidth}c}
\includegraphics[width=0.30\linewidth]{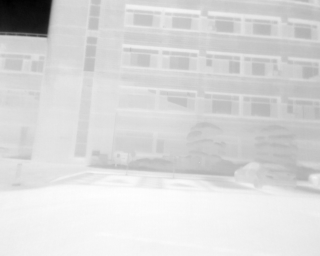} &
\includegraphics[width=0.30\linewidth]{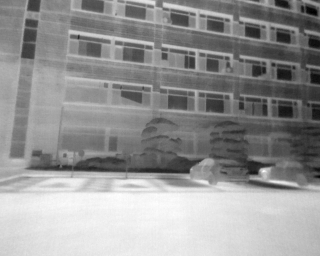} &
\includegraphics[width=0.30\linewidth]{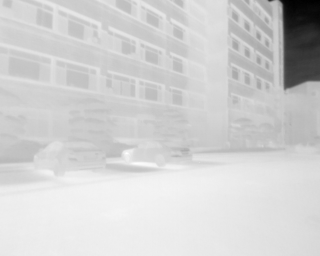} \\
{\footnotesize (a) Frame ${t-N}$} & {\footnotesize (b) Frame ${t}$ } & {\footnotesize (c) Frame ${t+N}$} \\
\end{tabular}
\end{center}
\caption{{\bf Problem analysis of normalized thermal images}. 
Typical thermal camera provides a normalized thermal image in default setting. 
Generally, the normalized thermal image is acquired through the thermal camera's image signal processing pipeline which normalizes the raw thermal image with its min-max values.
However, this process destroys temporal consistency, which is key-factor for self-supervised learning, between adjacent images.} 
\label{fig:temporal_incons}
\end{figure}

\section{Problem Analysis}
\label{sec:problem}
In this section, we conduct an in-depth analysis of thermal image characteristics that degenerate self-supervisory signals from thermal images. 

\subsection{Raw thermal image is suitable for temporal consistency}
The standard thermal camera can support two types of thermal images; raw thermal image and normalized thermal image.
Raw thermal image indicates thermal radiation values measured from thermal infrared detector array.
After that, thermal camera acquires a normalized thermal image by normalizing the raw thermal image with min-max values within raw measurement values.
However, the normalized thermal image loses the temporal consistency between adjacent images because of the different min-max value of each raw image, as shown in~\figref{fig:temporal_incons}.
Therefore, the raw thermal image is a suitable choice for the applications requiring self-supervision from temporal consistency. 
Shin~\etal~\cite{shin2021self} also partially discussed this topic and proposed a simple raw image processing strategy.
However, their method is insufficient to train depth and pose network solely, leading to adapt additional self-supervisory signals.

\begin{figure}[t]
\begin{center}
\footnotesize
\begin{tabular}
{c@{\hskip 0.005\linewidth}c@{\hskip 0.005\linewidth}c}
\includegraphics[width=0.301\linewidth]{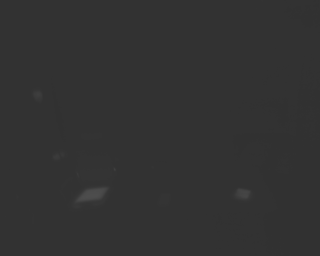} &
\includegraphics[width=0.30\linewidth]{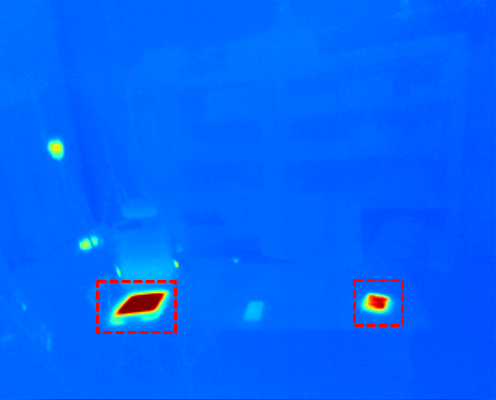} &
\includegraphics[width=0.28\linewidth]{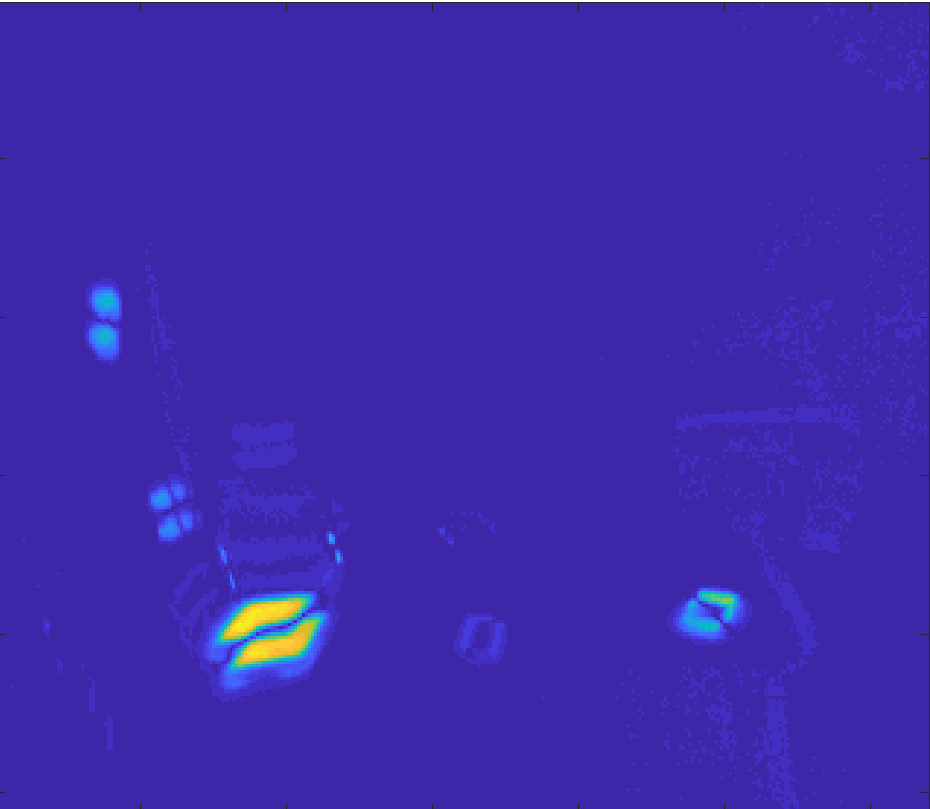} \\
{\footnotesize (a) Raw image} & {\footnotesize (b) Heat map } & {\footnotesize (c) Difference} \\
\multicolumn{3}{c}{\includegraphics[width=0.90\linewidth]{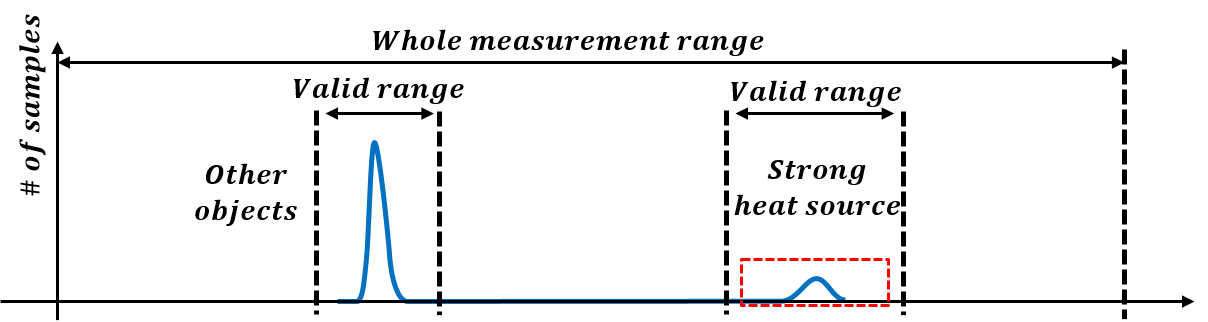}} \\
\multicolumn{3}{c}{{\footnotesize (d) Histogram of raw thermal image}} \\
\end{tabular}
\end{center}
\caption{{\bf Problem analysis of raw thermal image}. 
The typical thermal camera~\cite{flir-ax5} has an extensive dynamic range that can capture over than [-30\textdegree C, 150 \textdegree C].
From left to right, (a) is raw thermal image, (b) is heat map clipped with the range [10\textdegree C, 30\textdegree C], (c) is image difference map between adjacent raw images.
(d) is histogram plot of the raw thermal image.
The thermal image difference is dominated by an intense heat source and lost most of the object details' difference.} 
\label{fig:problem}
\end{figure}

\subsection{Sparsely distributed thermal radiation values dominate image reconstruction loss}
However, directly utilizing raw thermal images cannot generate a valuable self-supervisory signal from image reconstruction process.
Typical thermal camera~\cite{flir-ax5} has a wide dynamic range that can measure temperature values in a range of [-30\textdegree C, 150\textdegree C].
On the other hand, common real-world object temperatures are located in a narrow range around room and air temperatures.
Also, the histogram of raw thermal image usually possesses multiple temperature distributions, as shown in~\figref{fig:problem}-(d).
In both natural and man-made environments, every object emits different thermal radiation according to its thermal properties and its own temperature.
Some objects have high or low temperature values, such as the sun, electric devices, ice cups, and vehicles.
Most other objects have similar temperatures with room or air temperatures.
Therefore, the measured temperature distributions usually exist in a sparse way and have a narrow valid area. 

This phenomenon makes the raw thermal image and normalized image usually look to have weak contrast and details.
This tendency is more severe when a distinctly hot or cold temperature object exists within scene, as shown in~\figref{fig:problem}-(a) and (b). 
Therefore, as shown in~\figref{fig:problem}-(c), the difference map is dominated by an intense heat source, loses most of the object details, and provides awful self-supervision for depth and pose network training, 

To decrease this domination effect, raw thermal image needs to be rearranged with a valid range only.
However, the valid range of each temperature distribution can be dynamically changeable depending on the time, places, and temperature conditions. 
The rearrange process should capture a valid temperature range of each raw image and preserve temporal consistency between images regardless of surrounding ambient temperature conditions.

\subsection{High and low temperature objects degenerate self-supervision from thermal image}
Also, as shown in~\figref{fig:problem}-(b) and (d), most near-room-temperature objects exist in the narrow distribution with a large number of observations.
On the other hand, since thermal radiation from high-temperature object affects its local regions through heat reflection and transfer, high-temperature objects have a wide thermal radiation distribution with a small number of observations.
Therefore, even after capturing valid range only, the small number of observations from high-temperature objects takes a large portion of the range.

This property degenerates self-supervision from the thermal image.
As shown in~\figref{fig:problem}-(c), since local regions of high-temperature objects have large temperature variance, the difference map shows diverse error contrast, from yellow to blue scale, for its local regions.
On the other hand, the remaining region of the difference map shows low error contrast, from blue to not-visible color, despite its large number of observations.
However, the difference between global structures is more appreciated than local region differences to train depth networks for structure-aware depth estimation. 
Therefore, to maximize self-supervision from thermal images, the contrast between a large number of observations should be increased by adjusting each distribution range proportional to the number of samples.

\begin{figure}[t]
\begin{center}
\includegraphics[width=0.99\linewidth]{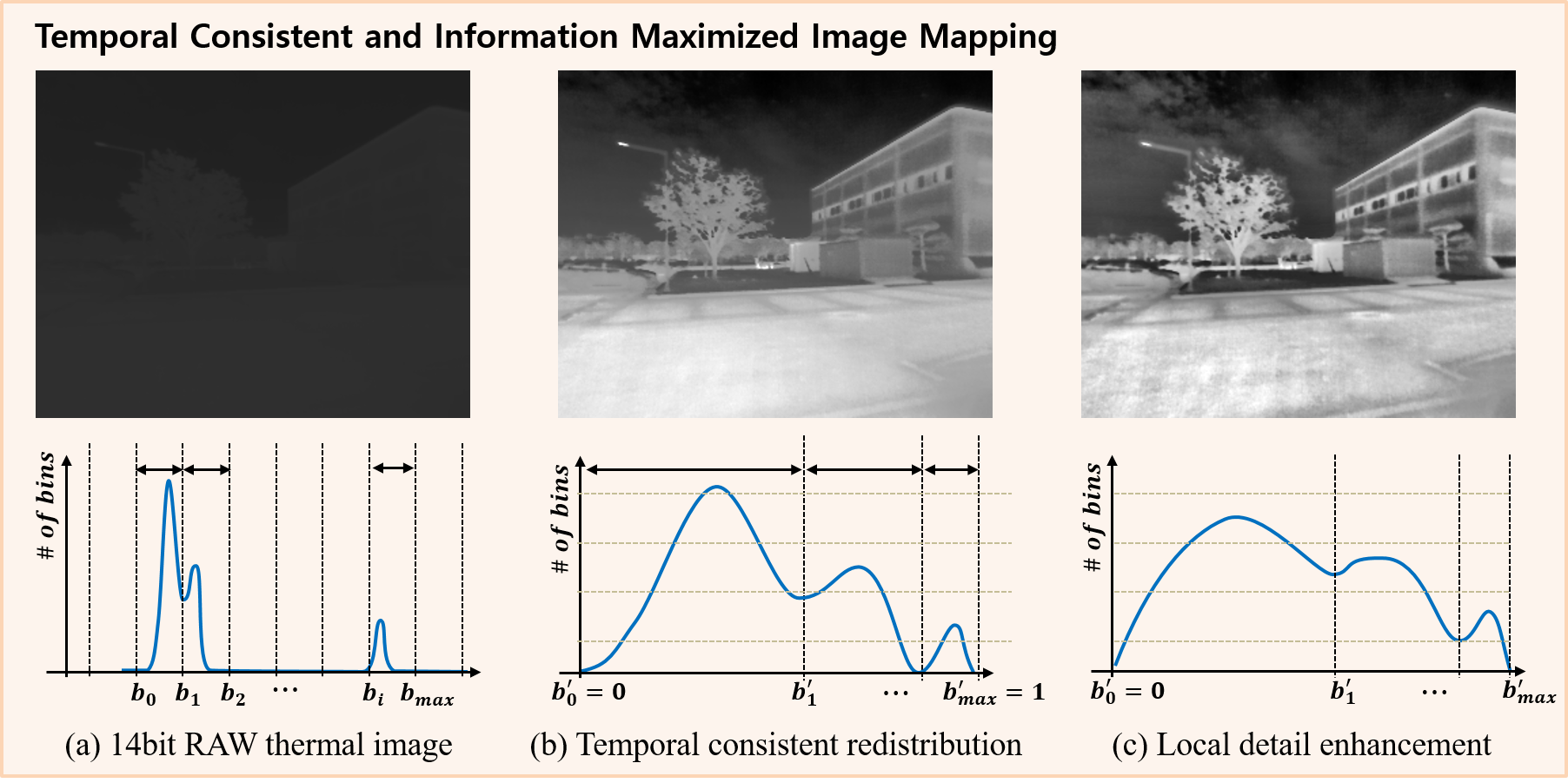} \\
\end{center}
\caption{{\bf Overall pipeline of temporal consistent and information maximized image mapping}. 
The proposed thermal image mapping process converts raw thermal images to have maximum image information while preserving temporal consistency locally.
From left to right, (a) is histogram of raw thermal images and corresponding one of raw images, (b) is rearranged histogram of one thermal image, (c) is locally enhanced thermal image.
After the mapping process, thermal image possesses greatly improved image information, such as overall structure, contrast, and details, by resolving domination and degeneration effects of raw thermal image.
}
\label{fig:TCIM}
\end{figure}

\section{Temporal Consistent and Information Maximized Image Mapping}
\label{sec:TCTR}
As discussed in~\secref{sec:problem}, we observed that the major degradation factors for self-supervision from thermal images are temporal inconsistency, sparsely distributed thermal radiation values, and few high- and low- temperature objects.
Therefore, to maximize self-supervision from thermal images while preserving temporal consistency, our intuitions are as follows:
\begin{itemize}
\item Rearranging thermal radiation value according to the observation number of each sub-histogram.
\item Maintaining temporal consistency by executing the rearrangement process in units of image groups to be used for loss calculation, not in units of single image.
\item
Enhancing local image details for a better self-supervisory signal.
\end{itemize}
The temporal consistent and information maximized image mapping process is shown in~\figref{fig:TCIM}.

\subsection{Temporal Consistent Thermal Radiation Rearrangement}

For given raw thermal images $I_{t},I_{s}$, the histogram $h(i)$ for raw temperature measurements is defined as: 
\begin{equation} 
\label{equ:histogram}
h(i)=n_i, \text{for } i = b_0, b_1, \cdots, b_{max},
\end{equation}
where $n_i$ is the number of samples within the range of $[b_{i}, b_{i+1})$ and $b_0,\cdots,b_i$ are calculated by dividing the global min-max values (\ie, $t_{min},t_{max}$) of raw images ($I_{t},I_{s}$) with a hyper-parameter bin number $N_{bin}$. 

After that, each raw measurement value $x$ within sub-histogram $[b_{i}, b_{i+1})$ are rearranged in proportional to its number of samples as follows: 
\begin{equation} 
\label{equ:rearrange}
x^{'} = \alpha_{i}*\left(\frac{x-b_{i}}{b_{i+1}-b_{i}} \right) +b_{i}^{'},
\end{equation}
where $\alpha_{i}$ is a scaling factor in proportional to the number of sub-histogram, defined as $\alpha_{i}=n_{i}/\sum_{j=0}^{max}{n_{j}}$. $b_{i}^{'}$ is a new offset of each scaled sub-histogram, defined as $b_{i}^{'}=\sum_{j=0}^{i-1}{\alpha_{j}}$.
This rearranging process stretches or squeezes each sub-histogram according to its number of samples and greatly increases image information, as shown in~\figref{fig:TCIM}-(b). 
Also, non-valid areas such as zero-observation regions are removed during this process. 
For a temporal consistency between adjacent images, the histogram, scaling factor, and offset calculation processes are conducted in units of image groups to be used for loss calculation, not in units of a single image.

\subsection{Local Detail Enhancement}
After thermal radiation rearrangement process, rearranged thermal image possesses greatly improved image information, such as overall structure, contrast, and details, by resolving domination and degeneration effects of raw thermal image.
However, we found that enhancing local image details can provide a better self-supervision from thermal image leading to performance improvements.
For this purpose, we adopt Contrast Limited Adaptive Histogram Equalization (CLAHE)~\cite{zuiderveld1994contrast} to avoid noise amplification and increase local image details while maintaining temporal consistency between adjacent frames as much as possible.
The CLAHE internally divide the image into non-overlapped local tiles, conduct per-tile histogram equalization with a pre-defined clip limit $\epsilon$.
Therefore, the rearranged thermal image $I^{'}$ is converted locally enhanced image $I^{eh}$, as follows:
\begin{equation} 
\label{equ:local_enhance}
I^{eh} = f_{LDE}(I^{'})
\end{equation}

\begin{figure}[t]
\begin{center}
\includegraphics[width=0.99\linewidth]{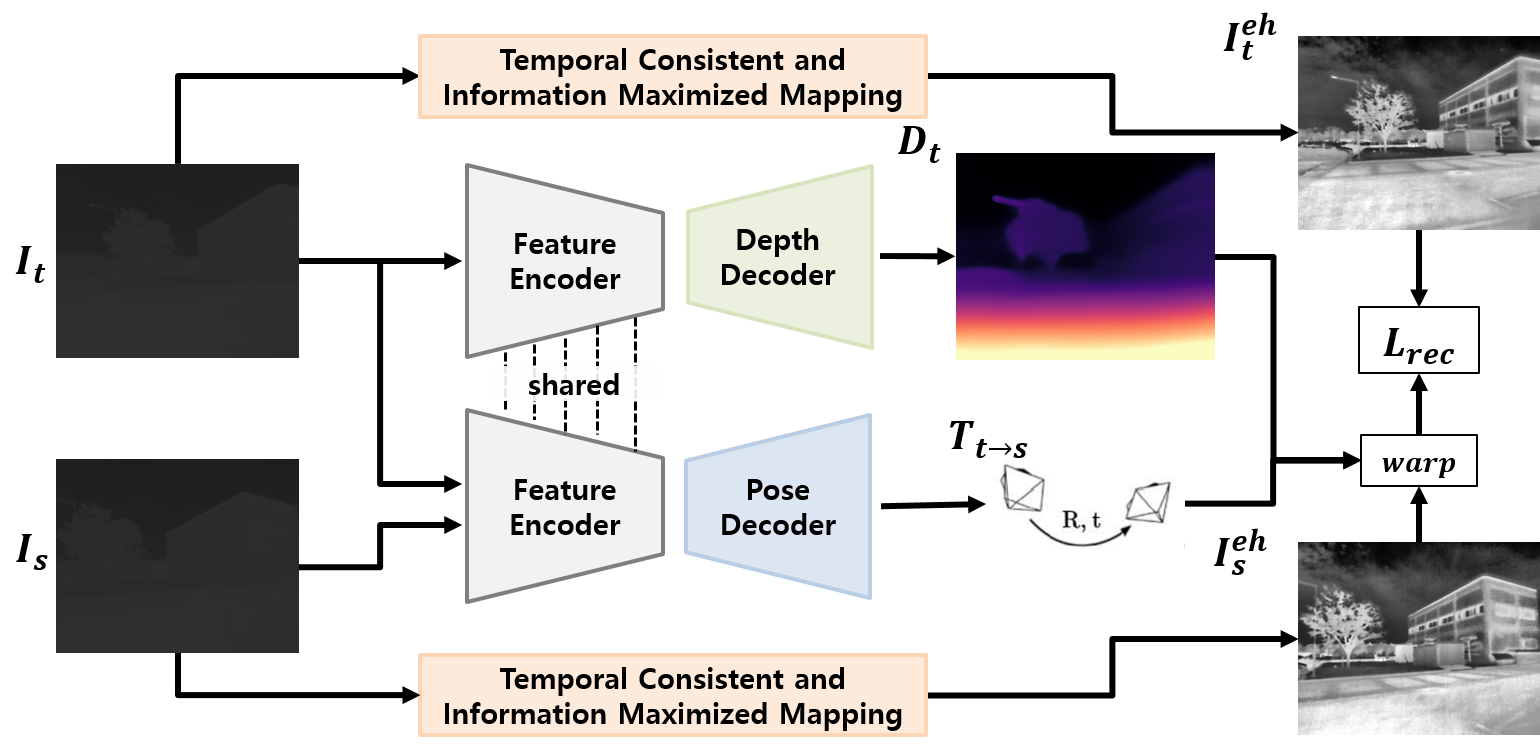} \\ 
\end{center}
\caption{{\bf Overall pipeline of our network architecture}. 
The depth and pose networks share the same feature encoder network and have task-specific decoder head.
The depth map $D_{t}$ and pose $T_{t\shortrightarrow{}s}$ are estimated from the raw thermal images $I_{t},I_{s}$ through the depth and pose networks. 
The temporal consistent temperature mapping module rearranges the raw images to have high contrast and object details (\ie, $I_{t}^{eh},I_{s}^{eh}$).
The thermal images are reconstructed with the estimated depth map $D_{t}$, relative pose $T_{t\shortrightarrow{}s}$, and source image $I_{s}^{eh}$.
}
\label{fig:overall_net}
\end{figure}

\section{Self-supervised Depth and Pose Learning form Monocular Thermal Video}
\label{sec:method}
In this section, we first describe network architecture for single-view depth and multiple-view pose estimation from thermal images.
After that, we describe overall and each self-supervised training loss to train depth and pose network from monocular thermal video. 

\subsection{Network architecture}
The depth and pose networks consist of a shared feature encoder and separate task-specific decoder heads, as shown in~\figref{fig:overall_net}. 
The shared feature encoder has ResNet18~\cite{he2016deep} backbone and is modified to take single-channel input thermal images.
We adopt the depth and pose decoder architecture of Monodepth v2~\cite{godard2019digging} for the depth and pose decoder heads.
The depth network takes a single-channel raw thermal image as an input and estimates a depth map.
The pose networks estimate a 6D relative pose from consecutive raw thermal images.

\subsection{Training Loss}
\label{sec:method_overview}
Our overall self-supervised training loss to train single-view depth and multiple-view pose estimation network is as follows:
\begin{equation} 
\label{equ:total loss}
\begin{split}
L_{total}&= L_{rec}+\lambda_{gc} L_{gc}+\lambda_{sm} L_{sm},
\end{split}
\end{equation}
where $L_{rec}$ indicates thermal image reconstruction loss, $L_{gc}$ is geometric consistency loss, $L_{sm}$ is edge-aware depth smoothness loss, and $\lambda_{gc}$ and $\lambda_{sm}$ are hyper parameters.
In the following descriptions, we use two consecutive images $[I_{t},I_{s}]$ (\ie, target and source images) for a concise explanation.
However, we utilizes three sequential images and both forward and backward direction loss in the training steps to maximize the data usage.

\subsubsection{Temperature Consistency Loss}
As shown in~\figref{fig:overall_net}, the depth and pose networks estimate a depth map $D_{t}$ and relative pose $T_{t\shortrightarrow{}s}$ from a raw thermal images $I_{t}, I_{s}$.
The raw thermal images are mapped to enhanced thermal images $I_{t}^{eh}, I_{s}^{eh}$ that possess high visibility, contrast, and details through the proposed mapping block.
After that, a synthesized thermal image $\Tilde{I}_{t}^{eh}$ is generated with the source image $I_{s}^{eh}$, target depth map $D_{t}$, and relative pose $T_{t\shortrightarrow{}s}$ in the inverse warping manner~\cite{zhou2017unsupervised}.
With the synthesized and original target thermal images, the thermal image reconstruction loss consisting of L1 difference and Structural Similarity Index Map (SSIM)~\cite{wang2004image} is calculated as follows:

\begin{equation} 
\label{equ:recon loss}
\begin{split}
L_{pe}(I_{t}^{eh}, \Tilde{I}_{t}^{eh}) = \frac{\gamma}{2}(1-SSIM(I_{t}^{eh}, \Tilde{I}_{t}^{eh})) \\ + (1-\gamma) ||I_{t}^{eh} - \Tilde{I}_{t}^{eh}||_{1},
\end{split}
\end{equation}
where $\gamma$ indicates scale factor between SSIM and L1 loss.

\subsubsection{Smoothness Loss}
As the temperature consistency loss usually does not provide informative self-supervision in low-texture and homogeneous regions, we regularize the estimated depth map to have smooth property by adding edge-aware smoothness loss $L_{sm}$~\cite{godard2019digging}.
\begin{equation} 
\label{equ:smothness loss}
L_{sm} = \sum{|\nabla D_{t}|\cdot e^{-|\nabla I_{t}^{eh}|}},
\end{equation}
where $\nabla$ is the first differential operator along spatial direction.

\subsubsection{Geometric Consistency Loss}
Geometric consistency loss $L_{gc}$~\cite{bian2019unsupervised} regularizes the estimated depth maps ($D_{t}$, $D_{s}$) to have scale-consistent 3D structure by minimizing geometric inconsistency.
The geometric inconsistency map $G_{diff}$ is defined as follows :
\begin{equation} 
\label{equ:geometric consistency}
G_{diff} = \frac{|\Tilde{D}_t-D_t^{'}|}{\Tilde{D}_t+D_t^{'}},
\end{equation}
where $\Tilde{D}_t$ is the synthesized depth map by warping the source depth map $D_{s}$ and relative pose $P_{t \shortrightarrow{} s}$.
$D_t^{'}$ is the interpolated depth map of $D_t$ to share the same coordinate with the synthesized depth map $\Tilde{D}_t$. 
After that, the geometry consistency loss is defined as follows :
\begin{equation} 
\label{equ:geometric consistency loss}
L_{gc} = \frac{1}{|V_p|}\sum_{V_p}{G_{diff}},
\end{equation}
where $V_{p}$ stands for valid points that are successfully projected from $I_{s}^{eh}$ to the image plane of $I_{t}^{eh}$ and $|V_{p}|$ defines the number of points in $V_{p}$.
During the training process, the scale consistency propagate to the entire video sequence. 

\subsubsection{Invalid Pixel Masking}
The per-pixel reconstruction loss includes invalid pixel differences caused by moving objects, static camera motion, and occlusion.
Therefore, we exclude the invalid reconstruction signal by checking depth consistency~\cite{bian2019unsupervised} and static pixel~\cite{godard2019digging}.
This pixel-wise reconstruction signal is filtered out according to the following~\eqnref{equ:recon mask loss}.
\begin{equation} 
\label{equ:recon mask loss}
\begin{split}
    L_{rec} &= \frac{1}{|V_{p}|}\sum_{V_{p}}{M_{gp}\cdot M_{sp}\cdot L_{pe}(I_{t}^{eh}, \Tilde{I}_{t}^{eh})},\\
    M_{gp} &= 1-G_{diff}, \\
    M_{sp} &= \left[L_{pe}(I_{t}^{eh}, \Tilde{I}_{t}^{eh})<L_{pe}(I_{t}^{eh}, I_{s}^{eh})\right], \\  
\end{split}
\end{equation}
where $M_{gp}$ is the geometric inconsistent pixel mask to exclude moving objects and occluded regions. 
Lastly, the static pixel mask $M_{sp}$ excludes the pixels which remains the same between adjacent frames because of a static camera motion and low texture regions. $[\cdot]$ is the Iverson bracket.



\section{Experimental Results}
\label{sec:result}

\subsection{Implementation Details}
\subsubsection{ViViD Dataset}
As described in~\cite{shin2021self}, there are very few dataset including the raw thermal images, Ground-Truth (GT) depth map, and GT pose labels.
Therefore, we also utilizes ViViD dataset~\cite{alee-2019-icra-ws} to validate effectiveness of the proposed method.
ViViD dataset consists of 10 indoor sequences and 4 outdoor sequence with different light and motion conditions.
We utilize 5 indoor sequences and 2 outdoor sequences as a training set and the remaining set as a test set. 

\subsubsection{Training setup}
We trained the depth and pose networks for the 200 epochs on the single RTX 2080Ti GPU with 12GB memory. 
Adam optimizer~\cite{kingma2014adam} is used to train the depth and pose networks.
The learning rates is set to $1e^{-6}$.
We utilize random crop and horizontal flip for the data augmentation of thermal images.
The input resolution of thermal image is 320-by-240.
We take about 16 hours to train the depth and pose networks.
The scale factors $\lambda_{gc}$ and $\lambda_{sm}$ are set to $0.5$ and $0.1$. 
The weight parameter $\gamma$ is set to $0.85$ for both indoor and outdoor datasets.
For the hyperparameter of temporal consistent temperature mapping module, we set bin number $N_{bin}$ as 30, CLAHE's clip threshold $\epsilon$ as 2 for outdoor set and 3 for indoor set, and CLAHE's local window size as $8$x$8$.

\begin{table*}[t]
\caption{\textbf{Quantitative comparison of depth estimation results on ViViD test set~\cite{alee-2019-icra-ws}}. 
We compare our network with state-of-the-art supervised/unsupervised depth networks~\cite{ranftl2020towards,bian2019unsupervised,shin2021self}. 
Input indicates the input source of each network. Supervision denotes the source of supervision (\eg, Depth is supervision from depth GT, and RGB or T is self-supervision from RGB or thermal video). 
Overall, $Ours$ shows comparable or outperformed results in all test set without additional modality guidance.
}
\begin{center}
\resizebox{0.95 \linewidth}{!}{
\def\arraystretch{1.3}
\footnotesize
\begin{tabular}{|c|c|c|c|c|c|cccc|ccc|}
\hline
\multirow{2}{*}{Scene} & \multirow{2}{*}{Methods} & \multirow{2}{*}{Architecture} & \multirow{2}{*}{Input} & \multirow{2}{*}{Supervision} &\multirow{2}{*}{Cap}  & \multicolumn{4}{c|}{\textbf{Error $\downarrow$}} & 
\multicolumn{3}{c|}{\textbf{Accuracy $\uparrow$}}    \\ \cline{7-13}
 & &  &  &  &  & AbsRel & SqRel & RMS & RMSlog &  $\delta < 1.25$ & $\delta < 1.25^{2}$ & $\delta < 1.25^{3}$ \\ 
\hline \hline
\multirow{7}{*}{\rotatebox{90}{\textbf{Indoor Well-lit}}}
    & Midas-v2~\cite{ranftl2020towards} & EfficientNet-Lite3 & RGB & Depth & 0-10m & 0.198 & 0.355 & 0.383 & 0.216 & 0.919 & 0.979 & 0.991 \\
    & Midas-v2~\cite{ranftl2020towards} & ResNext101 & RGB &  Depth & 0-10m & 0.194 & 0.348 & 0.370 & 0.210 & 0.928 & 0.979 & 0.991 \\
    & Midas-v2 & EfficientNet-Lite3  & Thermal &  Depth & 0-10m & 0.062 & 0.044 & 0.282 & 0.107 & 0.946 & 0.983 & 0.995 \\
    & Midas-v2 & ResNext101 &  Thermal &  Depth & 0-10m & \textbf{0.057} & \textbf{0.039} & \textbf{0.269} & \textbf{0.102} & \textbf{0.954} & \textbf{0.984} & \textbf{0.995} \\
    \cline{2-13}
    & SC-Sfm-Learner~\cite{bian2019unsupervised} & ResNet18 & RGB & RGB & 0-10m & 0.327  & 0.532 & 0.715 & 0.306 & 0.661 & 0.932 & 0.979 \\
    & Shin~\etal~\cite{shin2021self} (T)& ResNet18  & Thermal & Thermal & 0-10m & 0.225 & 0.201 & 0.709 & 0.262 & 0.620 & 0.920 & 0.993 \\
    & Shin~\etal~\cite{shin2021self} (MS)& ResNet18 & Thermal & RGB\&T & 0-10m & 0.156 & \textbf{0.111} & \textbf{0.527} & 0.197 & 0.783 & \textbf{0.975} & \textbf{0.997} \\
    & $Ours$ & ResNet18 & Thermal & Thermal & 0-10m & \textbf{0.152} & 0.121 & 0.538 & \textbf{0.196} & \textbf{0.814} & 0.965 & 0.992 \\
\hline
\hline
\multirow{7}{*}{\rotatebox{90}{\textbf{Indoor Dark}}}
    & Midas-v2~\cite{ranftl2020towards} & EfficientNet-Lite3   & RGB & Depth & 0-10m & 0.343 & 0.528 & 0.763 & 0.321 & 0.610 & 0.894 & 0.979 \\
    & Midas-v2~\cite{ranftl2020towards} & ResNext101 & RGB &  Depth & 0-10m & 0.351 & 0.545 & 0.766 & 0.327 & 0.624 & 0.875 & 0.976\\
    & Midas-v2 & EfficientNet-Lite3  & Thermal &  Depth & 0-10m & 0.060 & 0.036 & 0.273 & 0.105 & 0.950 & 0.985 & 0.996 \\
    & Midas-v2 & ResNext101 &  Thermal &  Depth & 0-10m & \textbf{0.053} & \textbf{0.032} & \textbf{0.257} & \textbf{0.099} & \textbf{0.958} & \textbf{0.987} & \textbf{0.996} \\
    \cline{2-13}
    & SC-Sfm-Learner~\cite{bian2019unsupervised} & ResNet18 & RGB & RGB & 0-10m & 0.452 & 0.803 & 0.979 & 0.399 & 0.493 & 0.786 & 0.933 \\
    & Shin~\etal~\cite{shin2021self} (T)& ResNet18 & Thermal & Thermal & 0-10m & 0.232 & 0.222 & 0.740 & 0.268 & 0.618 & 0.907 & 0.987 \\
    & Shin~\etal~\cite{shin2021self} (MS)& ResNet18 & Thermal & RGB\&T & 0-10m & 0.166 & 0.129 & 0.566 & 0.207 & 0.768 & 0.967 & 0.994 \\
    & $Ours$ & ResNet18 & Thermal & Thermal & 0-10m & \textbf{0.149} & \textbf{0.109} & \textbf{0.517} & \textbf{0.192} & \textbf{0.813} & \textbf{0.969} & \textbf{0.994} \\
\hline
\hline
\multirow{7}{*}{\rotatebox{90}{\textbf{Outdoor Night}}}
    & Midas-v2~\cite{ranftl2020towards} & EfficientNet-Lite3   & RGB     &  Depth & 0-80m & 0.278 & 2.382 & 7.203 & 0.318 & 0.560 & 0.821 & 0.946 \\
    & Midas-v2~\cite{ranftl2020towards} & ResNext101 & RGB     &  Depth & 0-80m & 0.264 & 2.187 & 7.110 & 0.306 & 0.571 & 0.833 & 0.955 \\
    & Midas-v2 & EfficientNet-Lite3   & Thermal &  Depth & 0-80m & 0.090 & 0.464 & 3.385 & 0.130 & 0.910 & 0.981 & 0.995 \\
    & Midas-v2 & ResNext101  & Thermal &  Depth & 0-80m & \textbf{0.078} & \textbf{0.369} & \textbf{3.014} & \textbf{0.118} & \textbf{0.933} & \textbf{0.988} & \textbf{0.996} \\
    \cline{2-13}
    & SC-Sfm-Learner~\cite{bian2019unsupervised} & ResNet18 & RGB & RGB & 0-80m & 0.617 & 9.971    & 12.000 & 0.595 & 0.400 & 0.587 & 0.720 \\
    & Shin~\etal~\cite{shin2021self} (T)& ResNet18 & Thermal & Thermal & 0-80m & 0.157	& 1.179	& 5.802	& 0.211	& 0.750	& 0.948	& 0.985 \\
    & Shin~\etal~\cite{shin2021self} (MS)& ResNet18 & Thermal & RGB\&T & 0-80m & 0.146	&  0.873	&  4.697	&  0.184	&  0.801	&  0.973	&  0.993 \\
    & $Ours$ & ResNet18 & Thermal & Thermal & 0-80m & \textbf{0.109} & \textbf{0.703} & \textbf{4.132} & \textbf{0.150} & \textbf{0.887} & \textbf{0.980} & \textbf{0.994} \\
\hline
\end{tabular}
}
\end{center}
\label{tab:depth result}
\end{table*}

\begin{table*}[t]
\caption{\textbf{Quantitative comparison of pose estimation results on ViViD dataset~\cite{alee-2019-icra-ws}}.
We compare our pose network with ORB-SLAM2~\cite{mur2017orb} and Shin~\etal~\cite{shin2021self}.
Note that ORB-SLAM2 often failed to track thermal image sequences. 
Therefore, the valid parts of the sequences that are successfully tracked are used for the accuracy calculation.
}
\begin{center}
\resizebox{0.95\linewidth}{!}
{
    \def\arraystretch{1.1}
    \footnotesize
    \begin{tabular}{|c|c|c||cc|cc|cc|}
    \hline
    \multirow{2}{*}{Scene} & \multirow{2}{*}{Methods} & \multirow{2}{*}{Input} & \multicolumn{2}{c|}{$M_{slow}$ + $I_{dark}$} & \multicolumn{2}{c|}{$M_{slow}$ + $I_{vary}$} & \multicolumn{2}{c|}{$M_{aggressive}$ + $I_{local}$} \\ 
    \cline{4-9} & & & ATE & RE & ATE & RE & ATE & RE \\ 
    \hline \hline
    \multirow{4}{*}{\rotatebox{90}{\textbf{Indoor}}}
        & ORB-SLAM2 & Thermal    & 0.0091$\pm$0.0066 & \textbf{0.0072$\pm$0.0035} & 0.0090$\pm$0.0088 & \textbf{0.0068$\pm$0.0034} & - & - \\
        & Shin~\etal~\cite{shin2021self} (T) & Thermal &  0.0063$\pm$0.0029 & 0.0092$\pm$0.0056 & 0.0067$\pm$0.0066 & 0.0095$\pm$0.0111 & \textbf{0.0225$\pm$0.0125} & 0.0671$\pm$0.055 \\
        & Shin~\etal~\cite{shin2021self} (MS) & Thermal & \textbf{0.0057$\pm$0.0030} & 0.0089$\pm$0.0050 & 0.0058$\pm$0.0032 & 0.0102$\pm$0.0124 & 0.0279$\pm$0.0166 & 0.0507$\pm$0.035 \\
        & $Ours$ & Thermal & 0.0059$\pm$0.0032 & 0.0082$\pm$0.0058 & \textbf{0.0054$\pm$0.0042} & 0.0083$\pm$0.0102 & 0.0293$\pm$0.0180 & \textbf{0.0489$\pm$0.0328} \\
    \hline
    \hline
    \multirow{2}{*}{Scene} & \multirow{2}{*}{Methods} & \multirow{2}{*}{Input} & \multicolumn{2}{c|}{$M_{slow}$ + $I_{night}$ (1)} & \multicolumn{2}{c|}{$M_{slow}$ + $I_{night}$ (2)} & & \\ 
    \cline{4-9} & & & ATE & RE & ATE & RE & & \\ 
    \hline
    \hline
    \multirow{4}{*}{\rotatebox{90}{\textbf{Outdoor}}}
        & ORB-SLAM2 & Thermal    & 0.1938$\pm$0.1380 & 0.0298$\pm$0.0150 & 0.1767$\pm$0.1094 & 0.0287$\pm$0.0126 & & \\
        & Shin~\etal~\cite{shin2021self} (T) & Thermal &  0.0571$\pm$0.0339 & \textbf{0.0280$\pm$0.0139} & 0.0534$\pm$0.029 & 0.0272$\pm$0.0121 & & \\
        & Shin~\etal~\cite{shin2021self} (MS) & Thermal &  0.0562$\pm$0.0310 & 0.0287$\pm$0.0144 & 0.0598$\pm$0.0316 & 0.0274$\pm$0.0124 & & \\
        & $Ours$ &  Thermal & \textbf{0.0546$\pm$0.0282} & 0.0287$\pm$0.0142 & \textbf{0.0509$\pm$0.0266} & \textbf{0.0271$\pm$0.0123} & & \\
    \hline
    \end{tabular}
}
\end{center}
\label{tab:ATE_result}
\end{table*}
\subsection{Single-view Depth Estimation Results}
We compare our proposed method with the state-of-the-art supervised/unsupervised depth networks~\cite{ranftl2020towards,bian2019unsupervised,shin2021self} to validate its effectiveness.
Note that the other related works~\cite{kim2018multispectral, lu2021alternative} require stereo RGB pair to train depth network. 
Therefore, their methods are not compatible with the ViViD dataset, including monocular RGB and thermal video only.
We train Midas-v2~\cite{ranftl2020towards} along with two architecture variants based on their pre-trained weight to investigate the upper bound.
We use evaluation metrics proposed by Eigen~\etal~\cite{eigen2014depth} to measure the performance of depth estimation results.
NYU~\cite{silberman2012indoor} and KITTI~\cite{geiger2013vision} evaluation settings are applied to indoor and outdoor set, respectively.

\begin{figure*}[t]
\begin{center}
\resizebox{0.98 \linewidth}{!}{
    \footnotesize
    \begin{tabular}{c@{\hskip 0.004\linewidth}c@{\hskip 0.004\linewidth}c@{\hskip 0.004\linewidth}c@{\hskip 0.004\linewidth}c@{\hskip 0.004\linewidth}c@{\hskip 0.004\linewidth}c@{\hskip 0.004\linewidth}}  
     \includegraphics[width=0.15\linewidth]{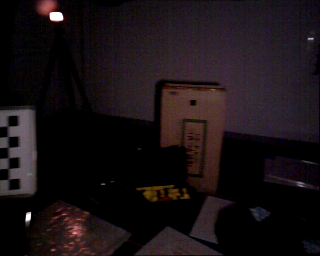} &
     \includegraphics[width=0.15\linewidth]{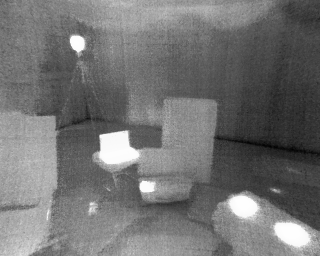} &
     \includegraphics[width=0.15\linewidth]{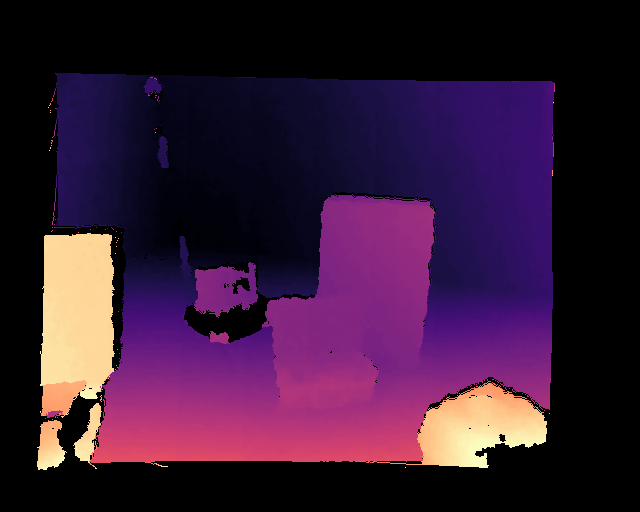} &
     \includegraphics[width=0.15\linewidth,height=2.12cm]{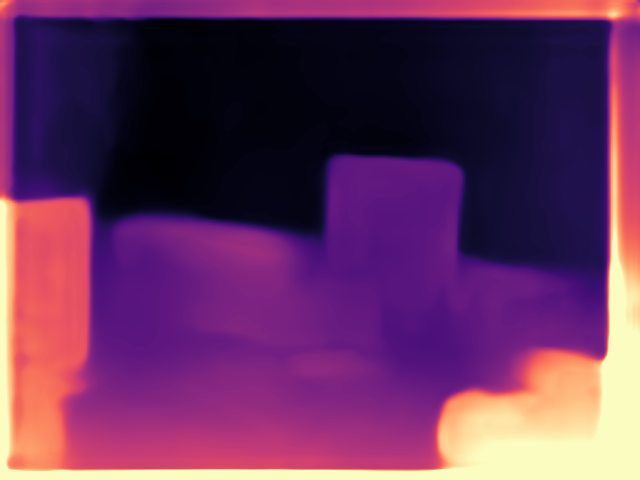} &
     \includegraphics[width=0.15\linewidth]{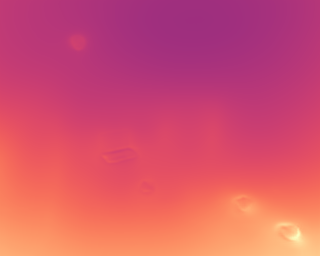} &
     \includegraphics[width=0.15\linewidth]{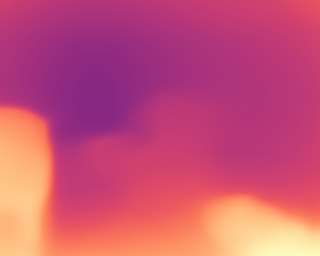} & 
     \includegraphics[width=0.15\linewidth]{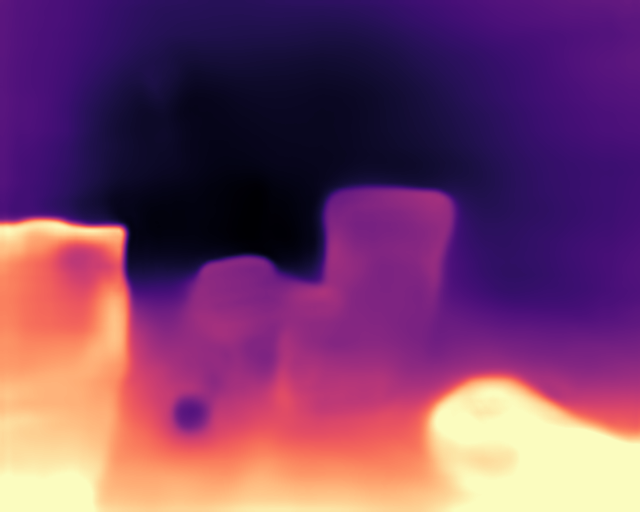} \vspace{-0.03in} \\
     \includegraphics[width=0.15\linewidth]{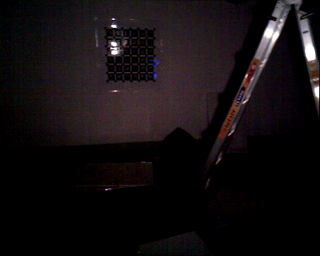} &
     \includegraphics[width=0.15\linewidth]{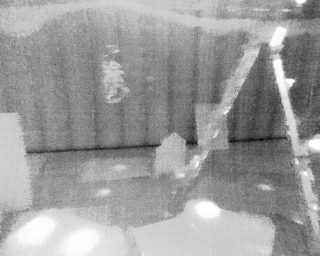} &
     \includegraphics[width=0.15\linewidth]{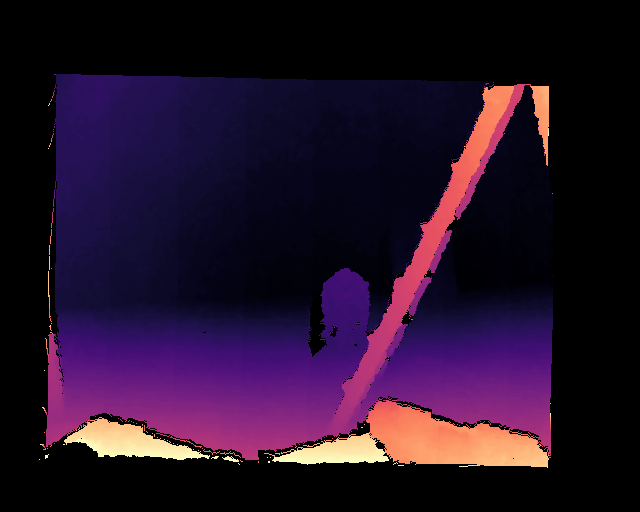} &
     \includegraphics[width=0.15\linewidth,height=2.12cm]{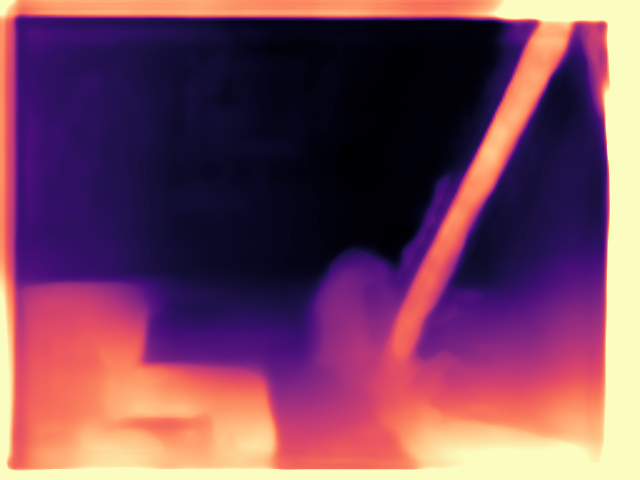} &
     \includegraphics[width=0.15\linewidth]{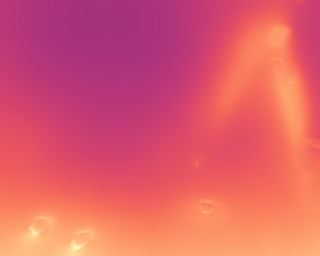} &
     \includegraphics[width=0.15\linewidth]{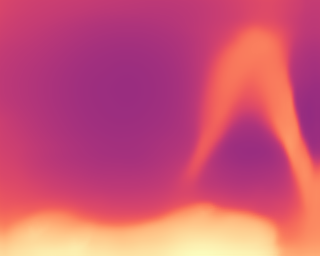} & 
     \includegraphics[width=0.15\linewidth]{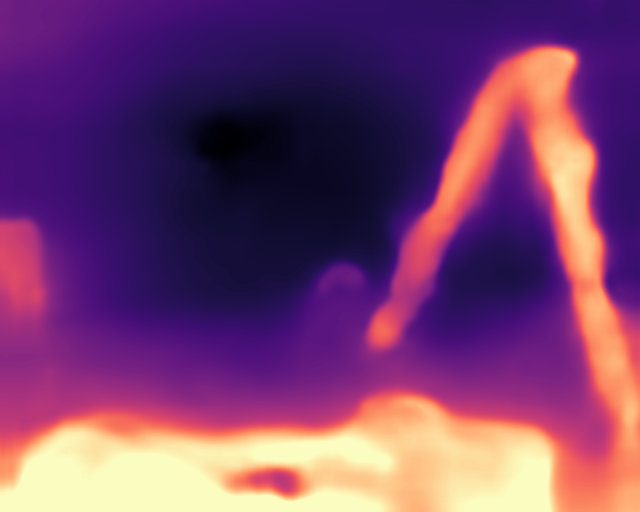} \vspace{-0.03in} \\
     \includegraphics[width=0.15\linewidth]{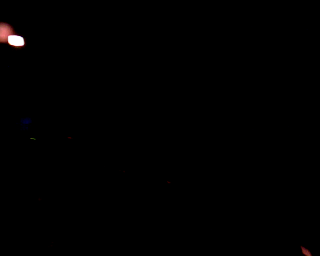} &
     \includegraphics[width=0.15\linewidth]{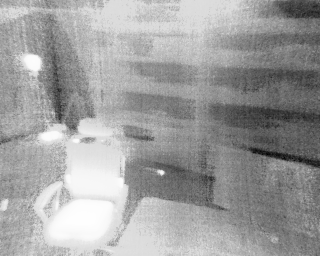} &
     \includegraphics[width=0.15\linewidth]{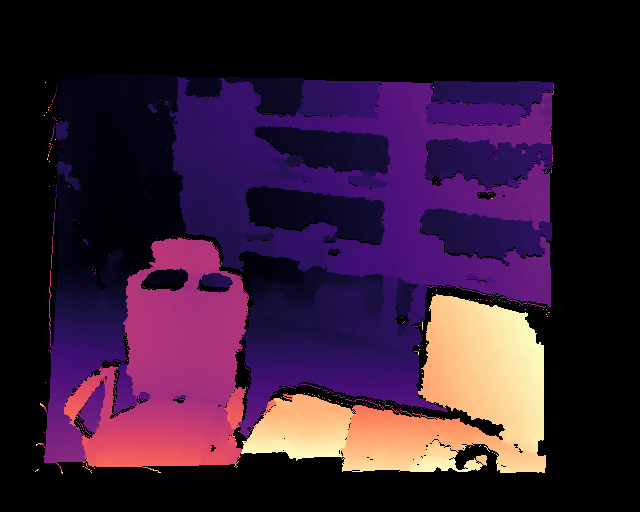} &
     \includegraphics[width=0.15\linewidth,height=2.12cm]{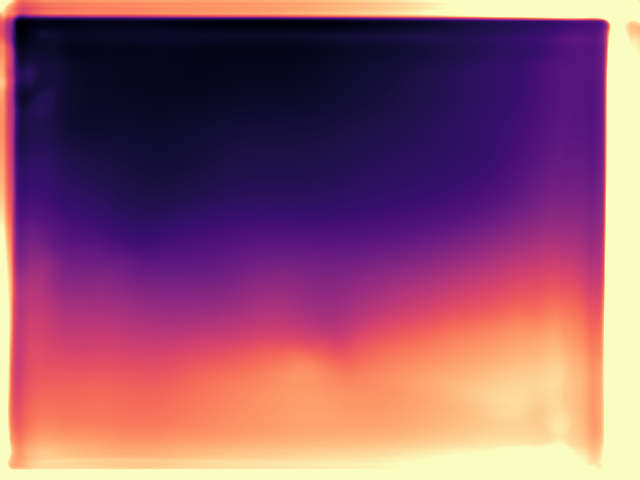} &
     \includegraphics[width=0.15\linewidth]{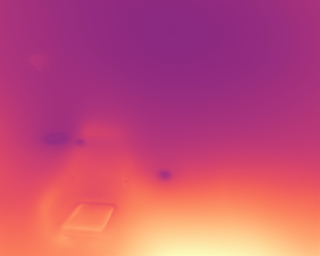} &
     \includegraphics[width=0.15\linewidth]{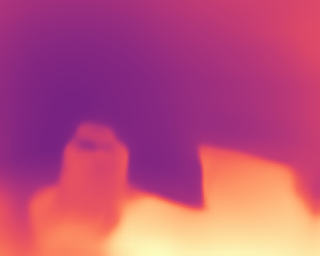} & 
    \includegraphics[width=0.15\linewidth]{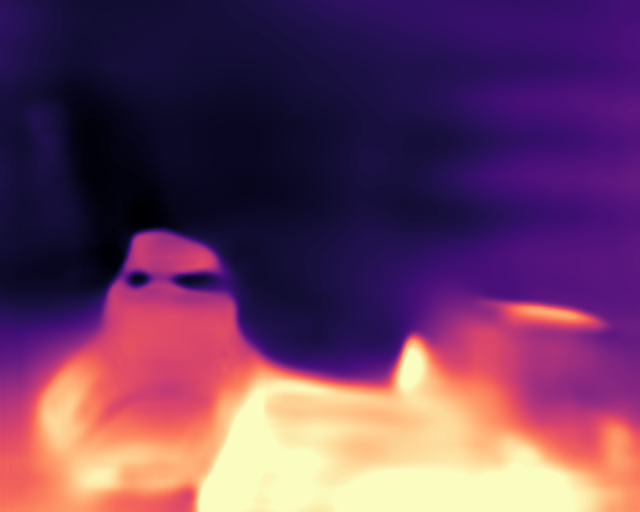} \vspace{-0.03in} \\
     \includegraphics[width=0.15\linewidth]{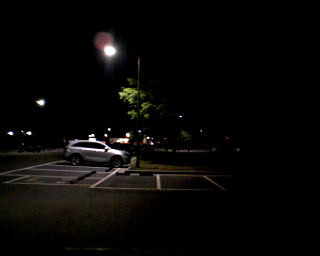} &
     \includegraphics[width=0.15\linewidth]{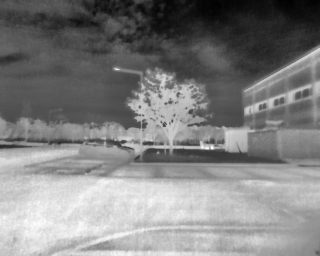} &
     \includegraphics[width=0.15\linewidth]{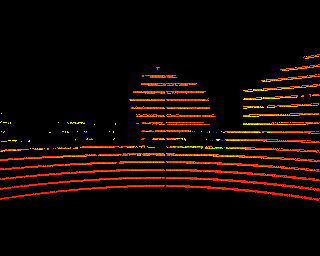} &
     \includegraphics[width=0.15\linewidth,height=2.12cm]{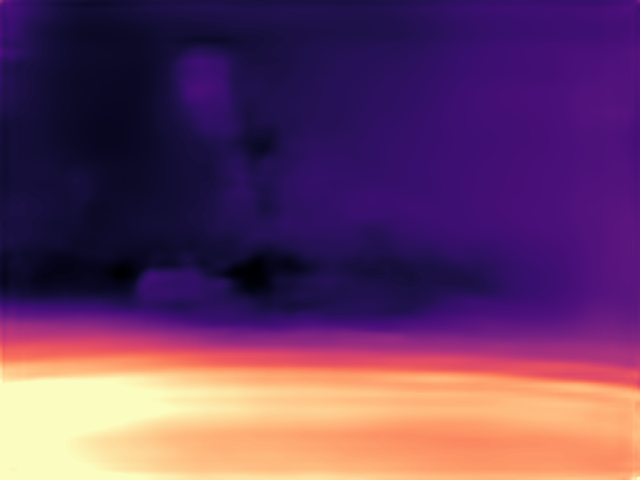} &
     \includegraphics[width=0.15\linewidth]{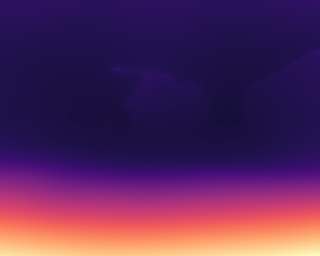} &
     \includegraphics[width=0.15\linewidth]{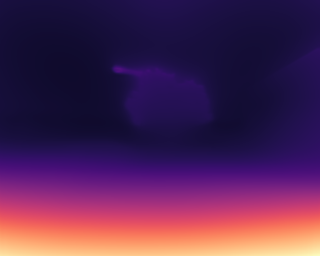} & 
    \includegraphics[width=0.15\linewidth]{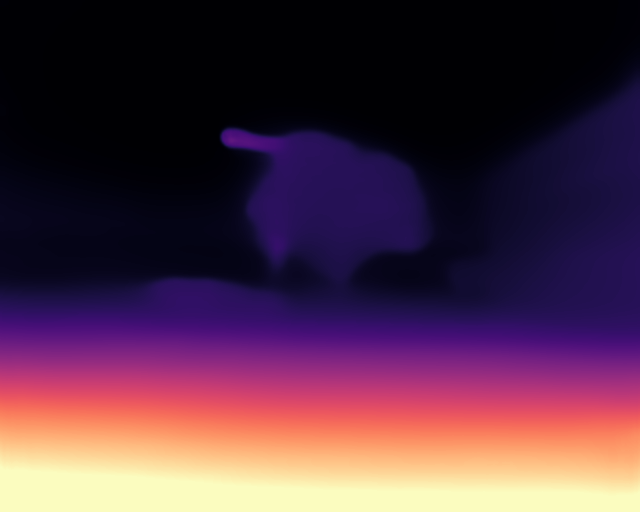} \vspace{-0.03in} \\
     \includegraphics[width=0.15\linewidth]{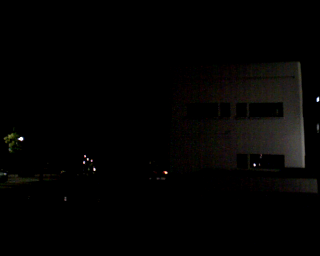} &
     \includegraphics[width=0.15\linewidth]{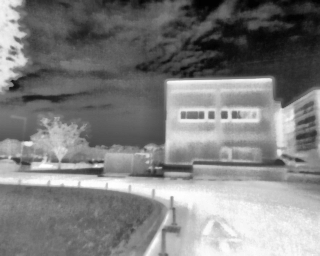} &
     \includegraphics[width=0.15\linewidth]{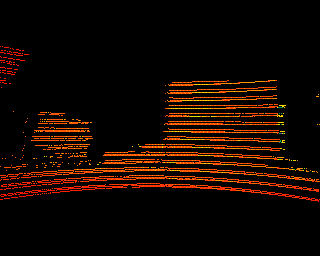} &
     \includegraphics[width=0.15\linewidth,height=2.12cm]{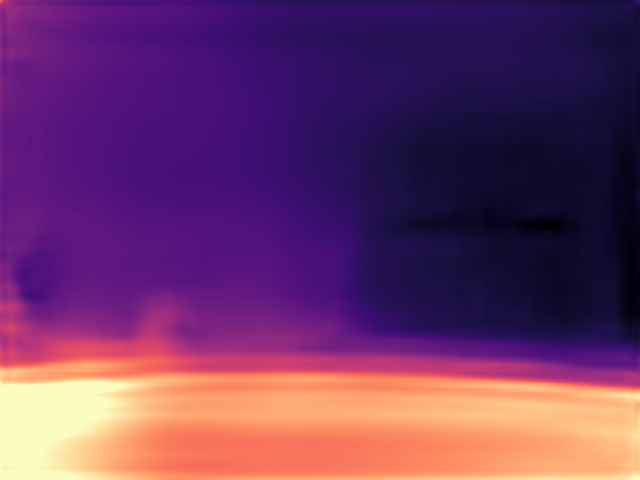} &
     \includegraphics[width=0.15\linewidth]{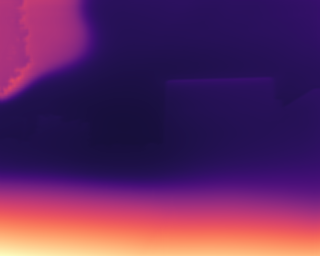} &
     \includegraphics[width=0.15\linewidth]{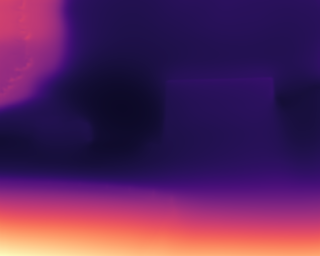} & 
    \includegraphics[width=0.15\linewidth]{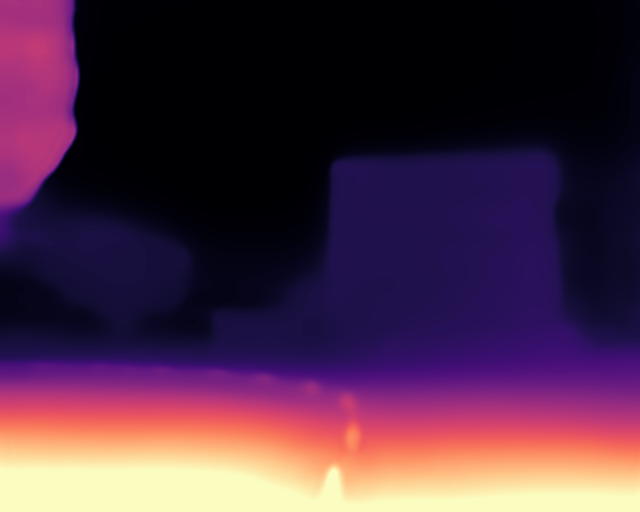} \vspace{-0.03in} \\
     \includegraphics[width=0.15\linewidth]{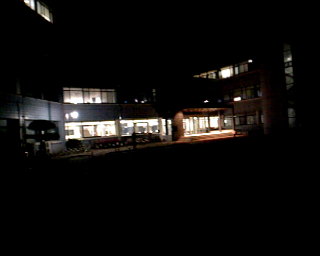} &
     \includegraphics[width=0.15\linewidth]{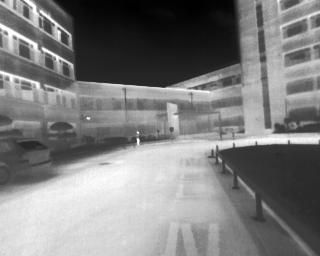} &
     \includegraphics[width=0.15\linewidth]{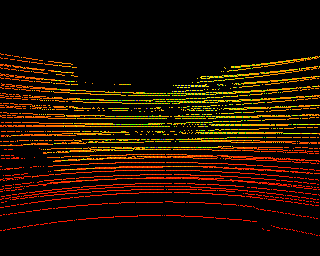} &
     \includegraphics[width=0.15\linewidth,height=2.12cm]{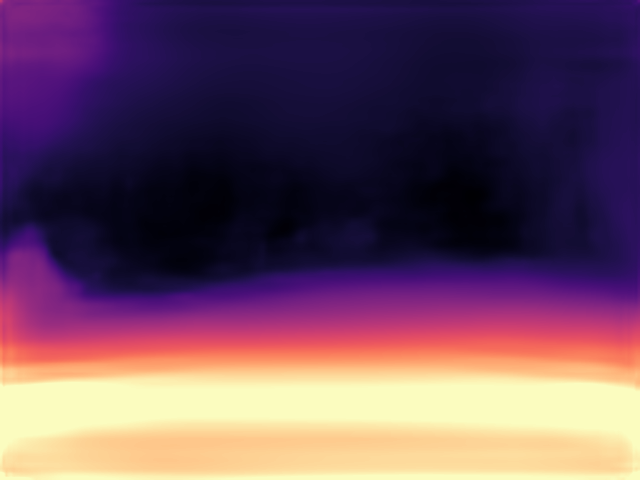} &
     \includegraphics[width=0.15\linewidth]{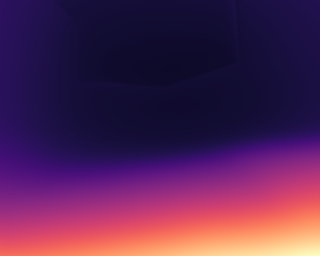} &
     \includegraphics[width=0.15\linewidth]{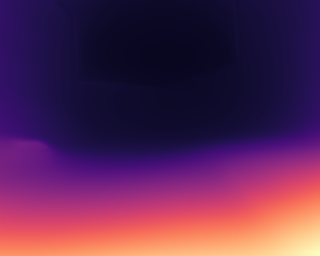} & 
     \includegraphics[width=0.15\linewidth]{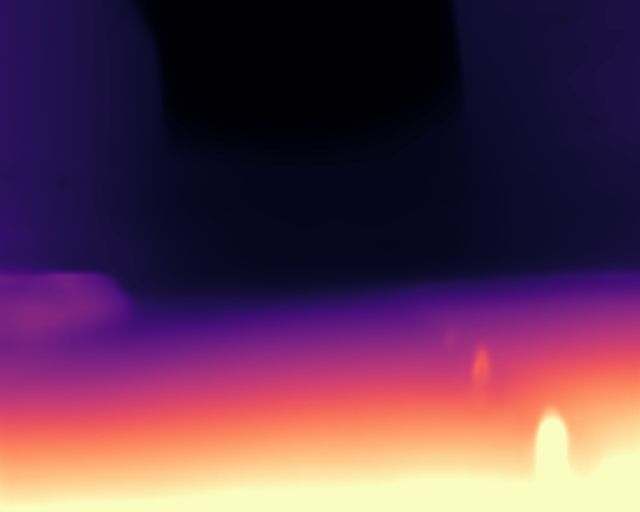} \vspace{-0.03in} \\
     {\footnotesize(a) Visible Image} & {\footnotesize (b) Thermal Image } & {\footnotesize (c) Ground Truth} & {\footnotesize (d) Midas-v2~\cite{ranftl2020towards}} & {\footnotesize (e) Shin-(T)~\cite{shin2021self}} & {\footnotesize (f) Shin-(MS)~\cite{shin2021self}} & {\footnotesize (g) $Ours$}\\
    \end{tabular}
    }
\end{center}
\caption{{\bf Qualitative comparison of depth estimation results on the ViViD dataset~\cite{alee-2019-icra-ws}}. 
Midas-v2 is EffcientNet based supervised depth network taking an RGB image.
The others are ResNet18 based self-supervised depth networks taking a thermal image.
The results show that the RGB image based network loses depth accuracy as the illumination condition gets worse.
On the other hand, thermal image based networks robustly estimate depth map results. 
Also, $Our$ shows sharp and accurate depth results without leveraging any additional modality guidance.
Note that, since the raw thermal image has low visibility, we instead visualize the processed thermal image with our method.
}
\label{fig:Exp_depth}
\end{figure*}

The experimental results are shown in~\tabref{tab:depth result} and~\figref{fig:Exp_depth}.
The RGB image based depth network provides precise depth results when the sufficient light conditions are guaranteed. 
However, as the light conditions worsen, the depth accuracy decreases significantly.
On the other hand, the thermal image based depth network shows generally robust depth estimation performance regardless of illumination condition.

Our proposed method $Ours$ shows outperformed performance than Shin~\etal~\cite{shin2021self} $(T)$ in all test datasets.
This fact indicates that our proposed method exploits self-supervision from monocular thermal video effectively than the previous method.
Also, $Ours$ greatly reduce the performance gap between supervised and self-supervised depth network in the outdoor set.
Shin~\etal~\cite{shin2021self} (MS) shows comparable results with $Ours$ in the indoor test set.
However, Shin~\etal~\cite{shin2021self} (MS) achieve this performance by relying on additional modality source, but our method exploits monocular thermal video only.
The reason for the limited performance improvement in the indoor set is discussed in~\secref{sec:discussion}.

\subsection{Pose Estimation Results}
We compare our pose estimation network with ORB-SLAM2~\cite{mur2017orb} and Shin~\etal~\cite{shin2021self}.
We utilize the 5-frame pose evaluation method~\cite{zhou2017unsupervised}. 
The evaluation metrics are Absolute Trajectory Error (ATE) and Relative Error (RE)~\cite{zhang2018tutorial}.
The experimental results are shown in~\tabref{tab:ATE_result}. 
The tendency is similar with the depth estimation results. 
Overall, $Ours$ shows better performance than Shin~\etal~\cite{shin2021self} $(T)$ in the both indoor/outdoor test sequences.
However, it shows comparable result with Shin~\etal~\cite{shin2021self} $(MS)$ in the indoor test sequencs.

\begin{table}[t]
 \centering
 \caption{\textbf{Ablation study on the temporal consistent temperature mapping method.} 
 The ablation studies are conducted in the outdoor test set of ViViD dataset.}
    \subfloat[\scriptsize The ablation study of sub-modules for depth estimation performance.]
    { 
        \resizebox{0.98\linewidth}{!}
        {
            \def\arraystretch{1.2}
            \begin{tabular}{c|cccc|ccc}
            \hline
            \multirow{2}{*}{Methods} & \multicolumn{4}{c|}{\textbf{Error $\downarrow$}} & 
            \multicolumn{3}{c}{\textbf{Accuracy $\uparrow$}}    \\ \cline{2-8}
             & AbsRel & SqRel & RMS & RMSlog &  $< 1.25$ & $< 1.25^{2}$ & $< 1.25^{3}$ \\ 
            \hline \hline
            $Base$    & 0.931	& 21.281 & 16.025	& 0.759	& 0.253	& 0.489	& 0.643 \\
            + TCTR  & 0.111  & 0.713  & 4.168  & 0.153  & 0.885  & 0.980  & 0.994 \\
            + LDE & \textbf{0.109} & \textbf{0.703} & \textbf{4.132} & \textbf{0.150} & \textbf{0.887} & \textbf{0.980} & \textbf{0.994} \\
            \hline
            \end{tabular}
            \label{table:abl_sub_mod}
        }
  }
  \hspace{1mm}
  \subfloat[\scriptsize Depth estimation performance according to the number of bin $N_{bin}$.]
    { 
        \resizebox{0.98\linewidth}{!}
        {
            \def\arraystretch{1.2}
            \begin{tabular}{c|cccc|ccc}
            \hline
            \multirow{2}{*}{Methods} & \multicolumn{4}{c|}{\textbf{Error $\downarrow$}} & 
            \multicolumn{3}{c}{\textbf{Accuracy $\uparrow$}}    \\ \cline{2-8}
             & AbsRel & SqRel & RMS & RMSlog &  $< 1.25$ & $< 1.25^{2}$ & $< 1.25^{3}$ \\ 
            \hline \hline
             $N_{bin}$=20         & 0.117 & 0.901 & 4.363 & 0.159 & 0.872 & 0.976 & 0.992  \\
             $N_{bin}$=30         & \textbf{0.109} & \textbf{0.703} & \textbf{4.132} & \textbf{0.150} & \textbf{0.887} & \textbf{0.980} & \textbf{0.994}  \\
             $N_{bin}$=40         & 0.112 & 0.811 & 4.228 & 0.155 & 0.884 & 0.979 & 0.993  \\
             \hline 
            \end{tabular}
            \label{table:abl_bin}
        }
  }
\captionsetup{font=footnotesize}
\label{table:Abl_all}
\end{table}

\subsection{Ablation Study}
\label{sec:ablation}
In this study, we investigate the effects of the proposed temporal consistent temperature mapping module, as shown in~\tabref{table:Abl_all}.
The $Base$ method uses an original raw thermal image to calculate temperature image reconstruction loss.
As discussed in~\secref{sec:problem}, strong heat sources dominate the raw thermal image, leading to insufficient self-supervision.
The Temporal Consistent Thermal radiation Rearranging (TCTR) module rearranges the temperature distribution in proportional to the number of samples within each sub-histogram. 
This process eliminates empty sub-histogram, decreases the effect of strong heat sources, and makes sufficient self-supervision for network training.
After that, the Local Detail Enhancement (LDE) further increases the performance by enhancing local details.
The hyperparameter $N_{bin}$ of TCTR module also investigated for its best performance. 

\begin{figure}[t]
\begin{center}
\footnotesize
\begin{tabular}
{c@{\hskip 0.05\linewidth}c}
\includegraphics[width=0.45\linewidth]{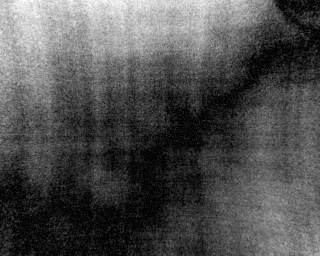} &
\includegraphics[width=0.45\linewidth]{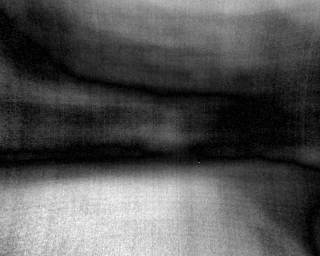} \\
{\footnotesize (a) Noise from indoor set} & {\footnotesize (b) Noise from outdoor set} \\
\end{tabular}
\end{center}
\caption{{\bf Low signal-to-noise ratio of thermal camera causes image noise}. 
From left to right, the averaged thermal image of whole indoor and outdoor sequences.
Since the thermal camera has low signal-to-noise ratio, the acquired images all include inevitable noise level to hinder effective self-supervision.
} 
\label{fig:dicsussion}
\end{figure}

\subsection{Discussion}
\label{sec:discussion}
We found the noise of thermal images is another inevitable problem for the effective thermal image reconstruction loss, as shown in~\figref{fig:dicsussion}.
In order to visualize thermal image noise existing over the whole dataset, we averaged the enhanced thermal images over the indoor and outdoor sequences.
Inherently, a thermal camera has a low signal-to-noise ratio property that is difficult to distinguish between noise and measurement value.
Also, any type of thermal image modification gradually increase image noise together.
This tendency is more severe when intense heat sources, such as the sun, don't exist (\eg, indoor).
Since the absence of a heat source makes most of the objects only have small temperature differences, increasing image contrast also amplifies image noise.
Therefore, the inevitable noise hinders self-supervised learning from thermal images, as shown in the indoor set result of~\tabref{tab:depth result}. 
Therefore, an effective noise handling method is another future research topic.

\section{Conclusion}
\label{sec:conclusion}
In this paper, we propose effective self-supervised learning method of depth and pose networks from monocular thermal video.
Based on the in-depth analysis for temporal thermal image reconstruction, we proposed the temporal consistent temperature mapping method that effectively maximizes image information while preserving temporal consistency.
The proposed method shows outperformed depth and pose results than previous state-of-the-art networks without leveraging additional RGB guidance.
We also discuss the remaining difficulty of the self-supervision from thermal images. 
We believe the proposed temporal consistent mapping method could be beneficial for other self-supervised learning tasks from thermal images.
Our source code is available at \url{https://github.com/UkcheolShin/ThermalMonoDepth}.


{
\bibliographystyle{IEEEtran}
\bibliography{IEEEexample}
}

\end{document}